\documentclass[runningheads]{llncs}
\usepackage{graphicx}
\usepackage{amsmath,amssymb} 
\usepackage{color}
\usepackage{caption}
\usepackage{subcaption}
\usepackage{multirow}
\usepackage{verbatim}
\usepackage{wrapfig,lipsum,booktabs}
\usepackage[width=122mm,left=12mm,paperwidth=146mm,height=193mm,top=12mm,paperheight=217mm]{geometry}

\DeclareMathOperator*{\argmax}{argmax}

\begin{document}
	\newcommand{\mat}[1]{\mathbf{#1}}
	\newcommand{\norm}[1]{\left\lVert#1\right\rVert}
\pagestyle{headings}
\mainmatter

\title{Query-Focused Extractive Video Summarization} 

\titlerunning{title running}

\authorrunning{}

\author{Aidean Sharghi, Boqing Gong, Mubarak Shah}


\institute{Center for Research in Computer Vision at UCF}

\maketitle

\begin{abstract}

Video data is explosively growing. As a result of the ``big video data", intelligent algorithms for automatic video summarization have (re-)emerged as a pressing need. We develop a probabilistic model, Sequential and Hierarchical Determinantal Point Process (\textsc{SH-DPP}), for {\bf query-focused} extractive video summarization. Given a {user query} and a long video sequence, our algorithm returns a summary by selecting key shots from the video. The decision to include a shot in the summary depends on the shot's relevance to the user query and importance in the context of the video, jointly. We verify our approach on two densely annotated video datasets. The query-focused video summarization is particularly useful for search engines, e.g., to display snippets of videos. 
\end{abstract}

\section{Introduction}
\label{Intro}
\begin{figure}[t]
\centering
		\includegraphics[width=\textwidth]{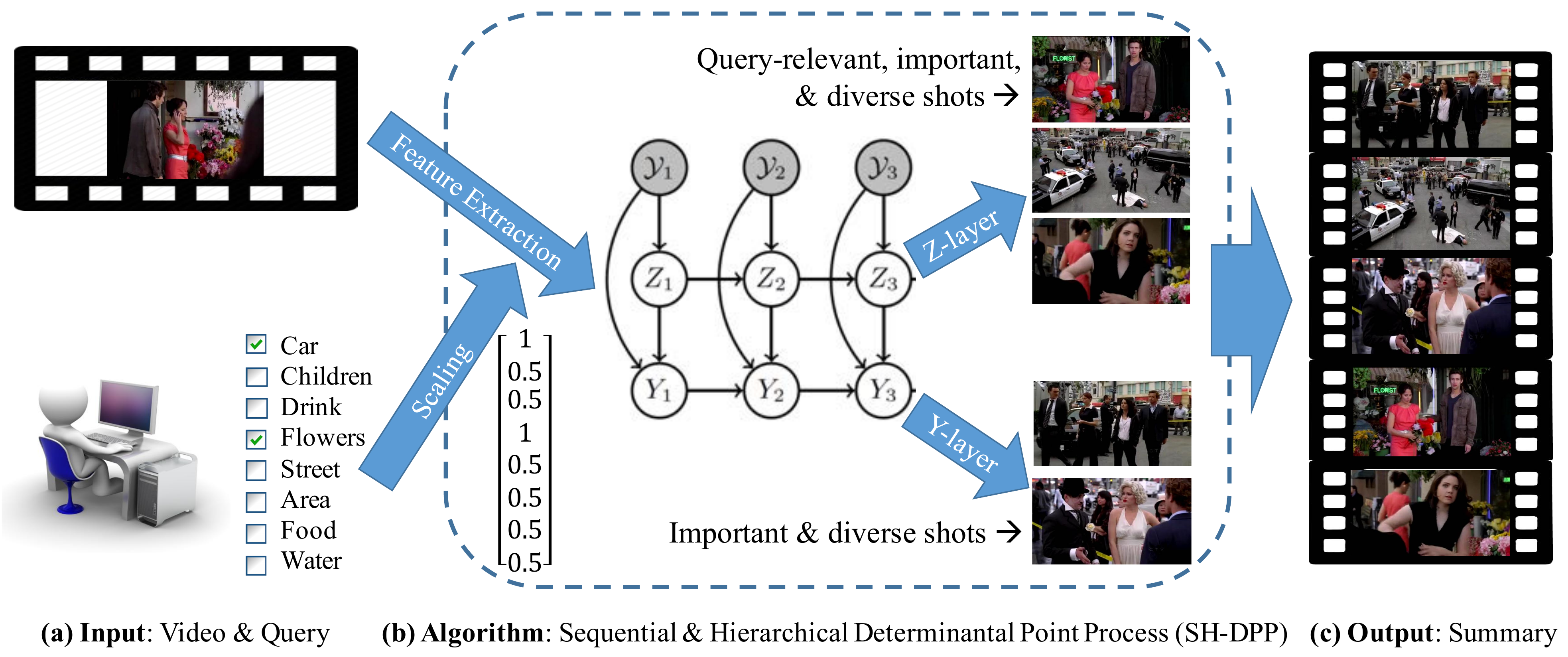}
	\caption{Query-focused video summarization and our approach to this problem. }
	\label{fig:fIllustrative}
	\vspace{-12pt}
\end{figure}

Video data is explosively growing as a result of the ubiquitous acquisition capabilities. The videos captured by UAVs and/or drones, from ground surveillance, and by body-worn cameras can easily reach the scale of gigabytes per day. There are about 30 million CCTV cameras perched around the United States surveying what is happening 24/7. About 300 hours of new videos are uploaded per minute to Youtube, let only many other online video sharing warehouses. While the ``big (video) data'' is a great source for information discovery and extraction, the computational challenges are unparalleled. In such context, intelligent algorithms for automatic video summarization (and retrieval, recognition, etc.) have thus (re-)emerged as a pressing need. 

In this paper we focus on \emph{extractive} video summarization, which generates a concise summary of a video by selecting from it the key frames or shots\footnote{It is perhaps more appealing to have a summary in the form of spatial-temporal synopsis or mosaic \emph{composed} of more than one frames. However, the \emph{compositional} video summarization is a highly challenging problem in general and has thus far achieved some success in only well-controlled environments~\cite{pritch2007webcam,pal2005interactive,kang2006space}.}. The key frames or shots are expected to be 1) {individually important}---otherwise they should not be selected, and 2) {collectively diverse}---otherwise one can remove some of them without losing much information. These two principles are employed in most of the existing works on extractive video summarization~\cite{jiang2009advances}, and yet implemented by different decision choices. Some earlier works define the importance of key frames by low-level appearance and/or motion cues~\cite{rav2006making,goldman2006schematic,liu2002optimization,aner2002video,vasconcelos1998spatiotemporal,wolf1996key}. 
Contextual information of a key frame is often modeled by graphs~\cite{lee2012unified,cong2012towards,ngo2003automatic}. We note that the system developers play a vital role in this cohort of works; most decisions on how to measure the importance and diversity  are handcrafted by the system developers using the low-level cues.

From the more recent works, we see a paradigm shift in some sort:  more high-level supervised information is introduced to video summarization than ever before. Rich Web images and videos provide (weakly) supervised priors for defining {\bf user-oriented} importance of the visual content in a video~\cite{khosla2013large,kim2014joint,xiong2014detecting,chu2015video}. For instance, the \textsc{car} images on the Web reveal the canonical views of the cars liked by average users, which should thus be given special attentions in video summarization. The texts associated with videos are undoubtedly good sources for inferring the semantic importance of video frames~\cite{song2015tvsum,liu2015multi}. Category-specific and domain-specific video summarization approaches are developed in~\cite{potapov2014category,sun2014ranking}. Some other high-level factors include gaze~\cite{xu2015gaze}, interestingness~\cite{gygli2014creating}, influence~\cite{lu2013story}, tracking of salient objects~\cite{lee2015predicting,liu2010hierarchical}, and so forth. 

What are the advantages of leveraging high-level supervised information in video summarization over merely low-level cues? We believe the main advantage is that the system developers are able to better infer the system {\bf users'} needs. After all, video summarization is a subjective process. Comparing to designing the system from the experts' own intuitions, it is more desirable to design a system based on the crowd or average users such that the system's states approach the users' internal states, which are often semantic and high-level.

What is the best supervision for a video summarization system? We have seen many types of supervision used in the above-mentioned works, such as Web images, texts, categories, etc. However, we argue that the best supervision, for the purpose of developing video summarization approaches, is the video summaries directly provided by the {\bf users/annotators}. In~\cite{gong2014diverse}, which is the first supervised video summarization work as far as we know, Gong et al.\ showed that there exists a high inter-annotator agreement in the summaries of the same videos given by distinct users. They proposed a supervised video summarization model, sequential determinantal point process (seqDPP), and train seqDPP  by the ``oracle'' summaries that agree the most among different user summaries. Gygli et al.\ gave another supervised approach to video summarization by learning submodular functions from the user summaries~\cite{gygli2015video}. 

From the low-level visual and motion cues to the high-level (indirect) supervised information, and to the  (direct) supervised user summaries, video summarization works become more and more {\bf user-oriented}. Though the two principles, importance and diversity, remain the same, the detailed implementation choices have significantly shifted from the system developers to the system users; users can essentially teach the system how to summarize videos in~\cite{gong2014diverse,gygli2015video}.

In respect to the recent progress, the goal of this paper is to further advance the user-oriented video summarization by modeling user input, or more precisely user intentions, in the summarization process. Figure \ref{fig:fIllustrative} illustrates our main idea. We name it query-focused (extractive) video summarization, in accordance with  the query-focused document summarization~\cite{nenkova2012survey} in  NLP. In this paper, a query refers to one or more concepts (e.g., \textsc{car, flowers}) that are both user-nameable and machine-detectable. More generic queries are left for the future work. 

Towards the goal of query-focused video summarization, we develop a probabilistic model, Sequential and Hierarchical Determinantal Point Process (\textsc{SH-DPP}). It has two layers of random variables, each of which serves for subset selection from a ground set of video shots (see Figure~\ref{fig:SOSeqDPP}). The first layer is mainly used to select the shots relevant to the user queries, and the second layer models the importance of the shots in the context of the videos. We condition the second layer on the first layer so that we can automatically balance the two strengths by learning from user labeled summaries. The determinantal point process (\textsc{DPP})~\cite{kulesza2012determinantal} is employed to account for the diversity of the summary.

A key feature in our work is that the decision to include a video shot in the summary is jointly dependent on the shot's relevance to the query and representativeness in the video. Instead of handcrafting any decision criteria, we use the \textsc{SH-DPP} probabilistic model to automatically learn from the user summaries (and the corresponding user queries and video sequences). In a sharp contrast to~\cite{gong2014diverse,gygli2015video} which model average users, our work closely tracks individual users' intentions from their input queries, and thus has greater potential to satisfy various user needs: distinct personal preferences (e.g., a patient user prefers more detailed and lengthy summaries than an impatient user), different interests over time even about the same video (e.g., a party versus a particular person in the party), etc. Finally, we note that our work is especially useful for search engines to produce snippets of videos.

Our main contribution is on the query-focused video summarization. Querying videos is not only an appealing functionality to the users but also an effective communication channel for the system to capture a user' intention. Besides, we develop a novel  probabilistic model, \textsc{SH-DPP}. Similarly to the sequential DPP (seqDPP)~\cite{gong2014diverse}, \textsc{SH-DPP} is efficient in modeling extremely lengthy videos and capable of producing summaries on the fly. Additionally, \textsc{SH-DPP} explicitly accounts for the user input queries. Extensive experiments on the UT Egocentric~\cite{ghosh2012discovering} and TV episodes~\cite{videoSET} datasets verify the effectiveness of \textsc{SH-DPP}. To our knowledge, our work is the first on query-focused video summarization.

\section{Related Work and Background}
\label{related}
In this section, we mainly discuss some related works on query-focused document summarization and some earlier works on interactive video summarization in the multimedia community. We will then describe some variations of \textsc{DPP} and contrast them to our \textsc{SH-DPP}. 
\vspace{-10pt}

\paragraph{{\bf Query-focused document summarization}} has been a long-standing track in the Text Retrieval Conference (\url{http://trec.nist.gov/}) and the Document Understanding Conference (DUC) (\url{http://duc.nist.gov/}). In DUC 2005 for example, participants were asked to summarize a cluster of documents given a user's query describing the information needs. Some representative approaches to this problem include \textsc{BayeSum}~\cite{daume2006bayesian}, \textsc{FastSum}~\cite{schilder2008fastsum}, and log-likelihood based method~\cite{gupta2007measuring} among others. Behind the vast research in this topic is the strong motivations by popular search engines and human-machine interactions. However, the counterpart in vision, query-focused video summarization, has not been well formulated yet. We make some preliminary efforts toward it through this work. 
\vspace{-10pt}

\paragraph{{\bf Interactive video summarization}} shares some spirits with our query-focused video summarization. The system in~\cite{ellouze2010s} allows users to interactively select some video shots to the summary while the system summarizes the remaining video. In contrast, in our system the users can use concept-based queries to influence the summaries without actually watching the videos. Besides, our approach is supervised and trained by user annotations, not handcrafted by the system developers. There are some other works involving users in the summarization related tasks, such as thumbnail selection~\cite{liu2015multi} and storyline-based video representation~\cite{xiong2015storyline}. Our work instead involves user input in the video summarization. 
\vspace{-10pt}

\paragraph{{\bf Determinantal point process (\textsc{DPP})}}\cite{kulesza2012determinantal} is employed in our SH-DPP to model the diversity in the desired video summaries. We give it a brief overview and also contrast SH-DPP to various DPP models. 

Denote by $\cal Y$ $= \{1,2,\dots,N\}$  the ground set. A (L-ensemble) DPP defines a discrete probability distribution over a subset selection variable $Y$,
\begin{align} 
P(Y=y) = {\det(\mat{L}_y)}/{\det(\mat{L}+\mat{I})}, \quad \forall y\subseteq \cal Y,  \label{eDef}
\end{align} 
where $\mat{I}$ is an identity matrix, $\mat{L} \in \mathbb{S}^{N \times N}$ is a positive semidefinite kernel matrix and is the distribution parameter, and $\mat{L}_y$ is a squared sub-matrix with rows and columns corresponding to the indices $y\subseteq \cal Y$. By default $\det({\mat{L}_\emptyset})=1$. 

DPP is good for modeling summarization because it integrates the two principles of individual importance and collective diversity. By the definition (eq.~(\ref{eDef})), the importance of an item is represented by $P(i\in Y)=\mat{K}_{ii}$ and the repulsion of any two items is captured by $P(i,j\in Y)=P(i\in Y)P(j\in Y)-\mat{K}_{ij}^2$, where $\mat{K}=\mat{L}(\mat{L}+\mat{I})^{-1}$. In other words, the model parameter $\mat{L}$ is sufficient to describe both the importance and diversity of the items being selected by $Y$. The readers are referred to Theorem 2.2 in~\cite{kulesza2012determinantal} for more derivations. 

A vanilla DPP gave rise to state-of-the-art performance on document summarization~\cite{kulesza2012learning,chaolarge}. Its variation, Markov DPP~\cite{affandi2012markov}, was used to maintain the diversity between multiple draws from the ground set. A sequential DPP (seqDPP)~\cite{gong2014diverse} was proposed for video summarization. Our SH-DPP brings a hierarchy to seqDPP and uses the first layer to take account of the user queries in the summarization (subset selection) process.

\vspace{-6pt}

\section{Approach}
\vspace{-4pt}
Our approach takes as input a user query $q$ (i.e., concepts) and a long video $\cal Y$, and outputs a query-focused short summary $y(q,\cal Y)$,
\begin{align}
y(q,{\cal Y})\leftarrow \argmax_{y\subseteq\mathcal{Y}} \; P(Y=y|q,{\cal Y}), \label{eInference}
\end{align}
which consists of some shots of the video. We desire {\bf four major  properties} from the distribution $P(Y=y|q,{\cal Y})$. i) It models the subset selection variable $Y$. ii) It promotes diversity among the items selected by $Y$. iii) It works efficiently given very long (e.g., egocentric) or even endlessly streaming (e.g., surveillance) videos. iv) It has some mechanism for accepting the user input $q$. Together, the properties motivate a Sequential and Hierarchical DPP (SH-DPP) as our implementation to $P(Y=y|q,{\cal Y})$. SH-DPP is built upon seqDPP~\cite{gong2014diverse}. Therefore, we firstly discuss some related methods---especially seqDPP,  how they meet some of the {\bf properties} but not all,  and then present the details of  SH-DPP.
\vspace{-8pt}

\subsection{Sequential DPP (seqDPP) with user queries}
In order to satisfy the first two {\bf properties} i) and ii), one can use a vanilla DPP (cf.\ eq.~(\ref{eDef})) to extract a diverse subset of shots as a video summary. Though this works well for multi-document summarization~\cite{kulesza2012learning}, it is unappealing in our context mainly due to two reasons. First, DPP sees the ground set (i.e., all shots in a video) as a bag, in which the permutation of the items has no effect on the output. In other words, the temporal flow of the video is totally ignored by DPP; it returns the same summary even if the shots are randomly shuffled. Second, the inference (eq.~(\ref{eInference})) cost is extremely high when the video is very long, no matter by exhaustive search among all possible subsets $y\subseteq {\cal Y}$ or greedy search~\cite{kulesza2012determinantal}. We note that the submodular functions also suffer from the same drawbacks~\cite{gygli2015video,xu2015gaze}. 

\begin{figure}[t]
\centering
\begin{tabular}{ccc}
		\includegraphics[width=0.47\textwidth]{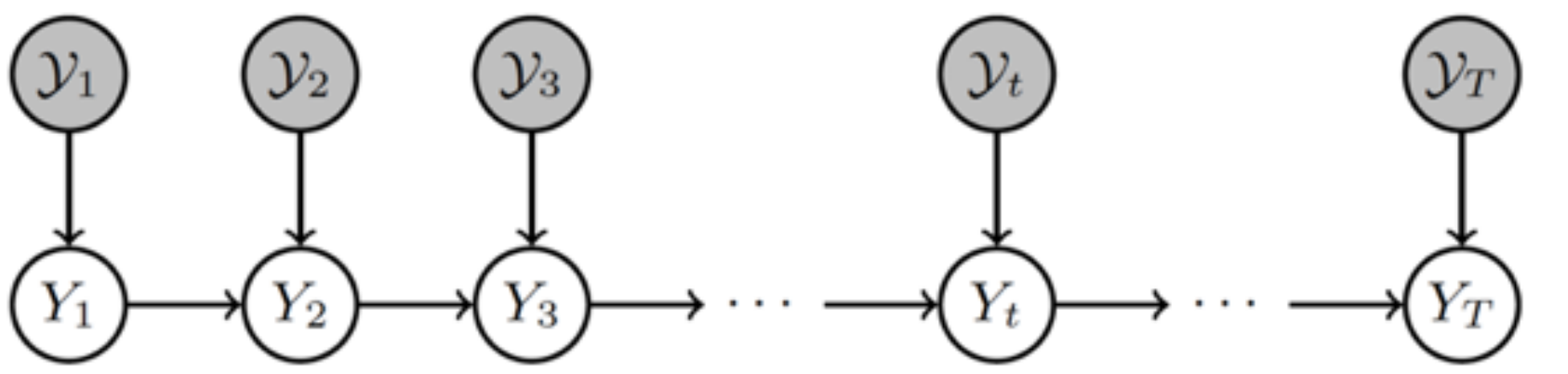} &~~~~~&
		\includegraphics[width=0.47\textwidth]{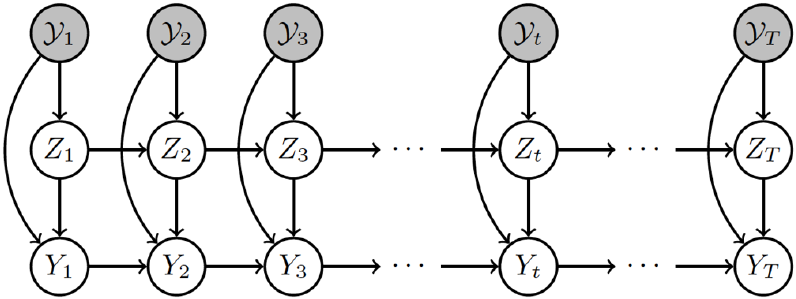}
\end{tabular}
	\caption{The graphical models of seqDPP~\cite{gong2014diverse} (left) and our SH-DPP (right).}
	\label{fig:SOSeqDPP}
	\vspace{-12pt}
\end{figure}

The seqDPP method~\cite{gong2014diverse} meets {\bf properties} i)--iii) and solves the problems described above. It partitions a video into $T$ consecutive disjoint sets, $\cup_{t=1}^T \cal Y$$_t=\cal Y$, where $\mathcal{Y}_t$ represents a set consisting of only a few shots and stands as the ground set of time step $t$. The model is defined as follows (see the left panel of Figure~\ref{fig:SOSeqDPP} for the graphical model),
\begin{align}
P_\textsc{seq}(Y|{\cal Y}) = P(Y_1|{\cal Y}_1)\prod_{t=2}^T P(Y_t|Y_{t-1}, {\cal Y}_t), \quad {\cal Y}=\cup_{t=1}^T {\cal Y}_t
\end{align}
where $P(Y_t |Y_{t-1}, {\cal Y}_t) \propto {\det \mat{\Omega}_{Y_{t-1}\cup Y_t}}$ is a conditional DPP to ensure diversity between the items selected at time step $t$ (by $Y_t$) and those of the previous time step (by $Y_{t-1}$). Similar to the vanilla DPP (cf.\ eq.~(\ref{eDef})), here the conditional DPP is also associated with a kernel matrix  $\mat{\Omega}$. In~\cite{gong2014diverse}, this matrix is parameterized by $\mat{\Omega}_{ij}=\vec{f}^T_iW^TW\vec{f}_j$, where $\vec{f}_i$ is a feature vector of the $i$-th video shot and $W$ is learned from the user summaries. Note that the seqDPP summarizer $P_\textsc{seq}(Y|{\cal Y})$ does not account for any user input. It is learned from ``oracle'' summaries in the hope of reaching a good compromise between distinct users. 

In this paper, we instead aim to infer individual users' intentional preferences over the video summaries, through the information conveyed by the user queries. To this end, a simple extension to seqDPP is to engineer query-dependent feature vectors $\vec{f}(q)$ of the video shots---details are postponed to Section~\ref{sFeature}. We consider this seqDPP variation as our baseline. It is indeed responsive to the queries through the query-dependent features, but it is fundamentally limited in modeling the query-relevant summaries, in which the importance of a video shot is jointly determined by its relevance to the query and how it is representative in the context of the video. The seqDPP offers no explicit treatment to the two types of interplayed strengths. In addition, the user may likely expect different levels of diversity from the relevant shots and irrelevant ones, respectively. However, a single DPP kernel in seqDPP fails to offer such flexibility. 

Our SH-DPP possesses all of the four {\bf properties}. It is developed upon seqDPP in order to take advantage of seqDPP's  nice properties i)--iii), and yet rectifies its downside (mainly on {\bf property} iv)) by a two-layer hierarchy. 
\vspace{-8pt}

%

\subsection{Sequential and Hierarchical DPP (SH-DPP)}
The right panel of Figure \ref{fig:SOSeqDPP} depicts the graphical model of SH-DPP, reading as,
\begin{align}
&P_\textsc{sh}(\{Y_1,Z_1\},...,\{Y_T,Z_T\}|\textcolor{red}{q},{\cal Y})  \notag \\
=& P(Z_1|\textcolor{red}{q},{\cal Y}_1) P(Y_1|Z_1,{\cal Y}_1)\prod_{t=2}^T P(Z_t|\textcolor{red}{q},Z_{t-1},{\cal Y}_t) P(Y_t|Z_t, Y_{t-1}, {\cal Y}_{t}).
\end{align}
Query $\textcolor{red}{q}$ is omitted from Figure \ref{fig:SOSeqDPP} for clarity. Each shaded node represents a video segment ${\cal Y}_t$ (i.e., a few consecutive and disjoint shots). We first use the subset selection variables $Z_t$ to select the query-relevant video shots. Note that $Z_t$ will choose zero shot if the segment ${\cal Y}_t$ does not contain any visual content related  to the query. Depending on the results of $Z_t$ (and $Y_{t-1}$), the variable $Y_t$ at the last layer selects video shots to further summarize the remaining content in the video segment ${\cal Y}_t$. The arrows in each layer impose diversity by DPP between the shots selected from two adjacent video segments---we thus have Markov diversities, in contrast to global diversity, in order to allow two (or more) similar shots to be simultaneously sampled to the summary if they appear at far-apart time steps (e.g., a man left home in the morning and returned home in the afternoon).

We define two types of  DPPs for the two layers of SH-DPP, respectively.
\vspace{-12pt}

%

\subsubsection{$Z$-Layer to summarize query-relevant shots.}
\label{subsec:SeqDPP}
Similar to seqDPP, we apply a conditional DPP $P(Z_t|\textcolor{red}{q},Z_{t-1},{\cal Y}_t)$ at each time step $t$ over the ground set ${\cal Y}_t \cup \{Z_{t-1}=z_{t-1}\}$, where ${\cal Y}_t$ consists of all the shots in partition $t$ and $z_{t-1}$ are the  shots selected by $Z_{t-1}$. In other words, the DPP here is conditioned on the selected items $z_{t-1}$ of the previous time step, enforcing diversity between two consecutive time steps,
\begin{equation}
P(Z_t = z_t|\textcolor{red}{q},Z_{t-1}=z_{t-1}, {\cal Y}_t) = \frac{\det \mat{\Omega}_{z_{t-1}\cup z_t}}{\det(\mat{\Omega}_{z_{t-1}\cup\mathcal{Y}_t} + I_t)}
\label{equ:Z-Layer}
\end{equation} 
where $I_t$ is the same as an identity matrix except that its diagonal values are zeros at the entries indexed by $z_{t-1}$. 

Different from seqDPP, we dedicate the $Z$-layer to query-relevant shots only. This is achieved by how  we train SH-DPP (Section~\ref{sTrain}) and the way we parameterize the DPP kernel matrix, 
\begin{align}
\mat{\Omega}_{ij}=[\vec{f}_i(\textcolor{red}{q})]^TW^TW[\vec{f}_j(\textcolor{red}{q})]  \label{eW}
\end{align}
where  $\vec{f}(\textcolor{red}{q})$ is a query-dependent feature vector of a shot. Details about the features are postponed to Section~\ref{sFeature}. In testing,  the  $Z$-layer only selects shots that are relevant to the user query $\textcolor{red}{q}$, and leaves all the unselected shots to the $Y$-layer which operates like a supplementary summarizer to the $Z$-layer.
\vspace{-12pt}


\subsubsection{$Y$-Layer to summarize the remaining video.} The decision to include a shot in the query-focused video summarization is driven by two interplayed forces. One is the shot's relevance to the query and the other is the representativeness of the shot. Given a user query $q$ (e.g., \textsc{car+flower}) and a long video ${\cal Y}$, likely many video shots are irrelevant to the query. As a result, we need another $Y$-layer to compensate the query-relevant shots selected by the $Z$-layer. In particular, we define the conditional probability distribution for the $Y$-layer variables as,
\begin{equation}
P(Y_t=y_t|Y_{t-1}=y_{t-1}, Z_{t}=z_t, \mathcal{Y}_t) = \frac{\det\mat{\Upsilon}_{y_{t-1}\cup z_t \cup y_t}}{\det(\mat{\Upsilon}_{y_{t-1}\cup \mathcal{Y}_t} + I_t')}
\label{equ:Y-Layer}
\end{equation}
where $y_{t-1}$ is the selected subset in previous time step at the $Y$-layer, $z_t$ is the selected subset of query-relevant shots in current time step by the $Z$-layer, and $I_t'$ is a diagonal matrix with ones indexed by $\mathcal{Y}_t\setminus z_t$ and zeros everywhere else.

Conditioning the $Y$-layer on the $Z$-layer has two advantages. First, no redundant information that is already selected by Z-layer is added by the $Y$-layer again to the summary, i.e., the shots selected by $Y$-layer are diverse from those by $Z$-layer. Second, $Y$-layer can, to some extent, compensate the missed query-relevant shots by $Z$-layer that were supposed to be selected. This occurs when such shots are themselves important in the context of the video.

Note that the $Y$-layer involves a new DPP kernel $\mat{\Upsilon}$, different from that used for the $Z$-layer. One obvious reason for doing this is that the two layers of variables serve to select different (query relevant or important) types of shots. Besides, it is also worth mentioning that the user may expect various levels of diversity from summary. When a user searches for \textsc{car+flower}, s/he probably would like to see some sort of redundancy in the shots of wedding car but not in the shots of police, making it necessary to have two types of DPP kernels. 

The $Y$-layer kernel is parameterized by 
\begin{align}
\mat{\Upsilon}_{ij} = \vec{f}_i^T\mat{V}^T\mat{V}\vec{f}_j  \label{eV}
\end{align}
and we will discuss how to extract features $\vec{f}$ from a shot in Section~\ref{sFeature}.
\vspace{-12pt}


\subsection{Training and testing SH-DPP}
\label{sTrain}
\vspace{-4pt}
The training data in our experiments are in the form of $(q,{\cal Y},z^q,y^q)$, where $z^q$ and $y^q$ respectively denote the query relevant and irrelevant shots in the summary. We learn the model parameters $\mat{W}$ and $\mat{V}$ of SH-DPP by maximum likelihood estimation (MLE): 
\begin{align}
\max_{\mat{W},\mat{V}} \quad \sum_q\sum_{\cal Y} \log P_\textsc{sh}(\{y_1,z_1\}, \cdots,\{y_T,z_T\}|q,\mathcal{Y}) - \lambda_1\|\mat{W}\|_F^2 - \lambda_2\|\mat{V}\|_F^2,    \label{eMLE}
\end{align}
where $\|\cdot\|_F^2$ is the squared Frobenius norm. We tune the hyper-parameters $\lambda_1$ and $\lambda_2$ by a leaving-one-video-out strategy, and optimize the above problem by gradient descent (cf.\ Supplement Material for more details on optimization).

After obtaining the local optimum $\mat{W}^*$ and $\mat{V}^*$ from the training, we need to know how to maximize the SH-DPP  $P_\textsc{sh}(y|q,\mathcal{Y})$ for the testing stage (cf.\ eq.~(\ref{eInference})). However, SH-DPP remains a computationally extensive combinatorial problem. We thus follow~\cite{gong2014diverse} to have an approximate online inference procedure: 
\begin{align}
{z_1}^* &= \argmax_{z \in \mathcal{Y}_1}{P(Z_1|\textcolor{red}{q},{\cal Y}_1)}, \qquad\;\;\;\,  {y_1}^* = \argmax_{y \in \mathcal{Y}_1\setminus z_1^*}{P(Y_1|z_1^*,{\cal Y}_1)}  \nonumber\\
{z_t}^* &= \argmax_{z \in \mathcal{Y}_t}{P(Z_t|\textcolor{red}{q},z_{t-1}^*,{\cal Y}_t)}, \quad  {y_t}^* = \argmax_{y \in \mathcal{Y}_t\setminus z_t^*}{P(Y_t|z_t^*, y_{t-1}^*, {\cal Y}_{t})}, \quad t\geq2  
\end{align}
where we exhaustively search for $z_t^*$ and $y_t^*$ from ${\cal Y}_t$ at each time step. Thanks to the online inference, we can readily use SH-DPP to handle very long  or even endlessly streaming videos. 


\vspace{-6pt}
\section{Experiment Setup}
\vspace{-4pt}
In this section, we describe the datasets, features of a video shot, user queries,  query-focused video summaries for training/evaluation, and finally, which metrics to  evaluate our learned video summarizer SH-DPP. 
\vspace{-8pt}

\subsection{Datasets}
\label{subsec:Dataset}
We use the UT Egocentric (UTE) dataset~\cite{ghosh2012discovering} and TV episodes~\cite{videoSET} whose dense user annotations are provided in~\cite{videoSET}. The UTE dataset includes four daily life egocentric videos, each 3--5 hours long, and the TV episodes contain four videos, each roughly 45 minutes long. These two datasets are very different in nature. The videos in UTE are long and recorded in an uncontrolled environment from the first-person view. As a result, many of the visual scenes are repetitive and likely unwanted in the user summaries. On the other hand, the TV videos are the professional episodes of TV series from the third person's viewpoint. All the scenes are well planned and controlled and mostly concise. A good summarization method should be able to work/learn well in both scenarios.

In \cite{videoSET}, all the UTE/TV videos are partitioned to 5/10-second shots, respectively, and for each shot a textual description is provided by a human subject. Additionally, for each video, 3 reference summaries are also provided each as a subset of the textual annotations. Thanks to the dense text annotations for every shot, we are able to derive from the text both user queries and two types of  query-focused video summaries, respectively, for patient and impatient users.


\begin{table*}[t]
\centering
\caption{\normalsize{The concepts  used in our experiments for the UTE and TV datasets.} }   \label{tQueries}
\begin{small}
\begin{tabular}{ll}
    \hline \toprule
    & \multicolumn{1}{c}{\normalsize{Concepts}} \\
    \hline 
      &area, band, bathroom, beach, bed, beer, blonde, boat, book, box, building, car\\
     & card, cars, chair, chest, children, chocolate, comics, cross, cup, desk, drink, eggs\\
     UTE & face, feet, flowers, food, friends, garden, girl, glass, glasses, grass, hair, hall\\
     & hands, hat, head, house, kids, lady, legs, lights, market, men, mirror, model \\
     & mushrooms, ocean, office, park, phone, road, room, school, shoes, sign, sky \\
     & street, student, sun, toy, toys, tree, trees, wall, water, window, windows (70)\\
    \hline
     & accident, animals, area, bat, beer, book, bugs, building, car, card, chair, child\\
    & clouds, commercial, cross, curves, dance, desk, drink, driver, evening, fan, father\\
    TV & food, girls, guy, hands, hat, head, history, home, house, kids, leaves, men, model\\
    & mother, office, painting, paintings, parents, phone, places, present, room, scene\\
    & space, storm, street, team, vehicle, wonder (52 in total)\\
    \bottomrule
\end{tabular}
\end{small}
\vspace{-16pt}
\end{table*}

\vspace{-6pt}

\subsection{User Queries}
\label{sQuery}
In this paper, a user query comprises one or more noun concepts (e.g., \textsc{car, flower, kid}); more generic queries are left for future research. There are many nouns in the text annotations of the video shots, but are they all useful for users to construct queries? Likely no. Any useful nouns have to be machine-detectable so that the system can ``understand'' the user queries. To this end, we construct a lexicon of concepts by overlapping all the nouns in the annotations with the nouns in SentiBank~\cite{borth2013sentibank}, which is a large collection of visual concepts and corresponding detectors. This results in a lexicon of 70/52 concepts for the UTE/TV dataset (see Table~\ref{tQueries}). Each pair of concepts is considered as a user query for both training and testing our SH-DPP video summarizer. Besides, at the testing phase, we also examine novel queries---all the triples of concepts. 

\vspace{-6pt}

\subsection{Query-Focused Video Summaries }
For each input query and video, we need to know the ``groundtruth'' video summary for training and evaluating SH-DPP. We construct such summaries based on the ``oracle'' summaries introduced in~\cite{gong2014diverse}. 

\subsubsection{Oracle summary.} As mentioned earlier in Section \ref{subsec:Dataset}, there are three human-annotated summaries $y^u, u=1,2,3$ for each video ${\cal Y}$. An ``oracle'' summary $y^o$ has the maximum agreement with all of the three annotated summaries, and can be understood as the summary by an ``average'' user. Such a summary is found by a greedy algorithm~\cite{kulesza2012learning}. Initialize $y^o=\emptyset$. In each iteration, the set $y^o$ increases by one video shot $i$ which gives rise to the largest marginal gain $G(i)$,
\begin{equation}
y^o \leftarrow y^o \cup \argmax_{i\in{\cal Y}} G(i), \quad G(i)=\sum_{u}\text{F-score}({y^o \cup i, {y}_u}) - \text{F-score}({y^o, {y}_u})
\end{equation}
where the F-score follows~\cite{videoSET} and is explained in Section~\ref{sEvaluation}. The algorithm stops when there is no such shot that the  gain $G(i)$ is greater than 0. Note that thus far the oracle summary is independent of the user query.
\vspace{-12pt}

\subsubsection{Query-focused video summary.} 
We consider two types of users. A {\bf patient user} would like to watch all the shots relevant to the query in addition to the summary of the other visual content of the video. For example, all the shots whose textual descriptions have the word \textsc{car} should be included in the summary if \textsc{car} shows up in the query. We union such shots with the oracle summary to have the query-focused summary for the patient user. On the other extreme, an {\bf impatient user} may only want to check the existence of the relevant shots, in contrast to watching all of them. To conduct experiments for the impatient users, we overlap the concepts in the {oracle summary} with the concept lexicon (cf.\ Section~\ref{sQuery}), and generate all possible bi-concept queries from the survived concepts. Note that the oracle summaries are thus the gold standards for training video summarizers for the impatient users.



\vspace{-6pt}

\subsection{Features}
\label{sFeature}

We extract high-level concept-oriented features $\vec{h}$ and contextual features $\vec{l}$ for a video shot. For each concept in the lexicon  (of size 70/52 for the UTE/TV dataset), we firstly use its corresponding SentiBank detector(s)~\cite{borth2013sentibank} to obtain the detection scores of the key frames, and then average them within each shot. Some of the concepts each maps to more than one detectors. For instance, there are beautiful \textsc{sky}, clear \textsc{sky}, and sunny \textsc{sky} detectors for the concept \textsc{sky}. We max-pool their shot-level scores, so there is always one detection score, which is between 0 and 1, for each concept. The resultant high-level concept-oriented feature vector $\vec{h}$ is 70D/52D for a shot of a UTE/TV video. We $\ell_2$ normalize it.

Furthermore, we design some contextual features $\vec{l}$ for a video shot based on the low-level features that SentiBank uses as input to its classifiers. This set of low-level features includes: color histogram, GIST~\cite{oliva2001modeling}, LBP~\cite{ojala2002multiresolution}, Bag-of-Words descriptor, and an attribute feature~\cite{yu2013designing}. With these features, we put a temporal window around each frame, and compute the mean-correlation as a contextual feature for the frame. The mean-correlation shows how well the frame is representative of the other frames in the temporal window. By varying the window size from 5 to 15 with step size 2, we obtain a 6D feature vector. Again we average pool  them within each shot, followed by $\ell_2$ normalization, to have the shot-level contextual feature vector $\vec{l}$. 

The concept-oriented and contextual features are concatenated as the overall shot-level feature vector~${\vec{f}}\equiv[\vec{h};\vec{l}]$ for parameterizing the DPP kernel of the $Y$-layer (eq.~(\ref{eV})). The $Z$-layer kernel calls for query-dependent features $\vec{f}(q)$ (eq.~(\ref{eW})). For this purpose, we scale the concept-oriented features according to the query: $\vec{f}(q)\equiv\vec{h}\circ \vec{\alpha}(q)$, where $\circ$ is the element-wise product between two vectors, and the scaling factors $\vec{\alpha}(q)$ are 1 for the concepts shown in the query and 0.5 otherwise (see Figure~\ref{fig:fIllustrative}(a, b) for an example).  Though we may employ more sophisticated query-dependent features, the simple features scaled by the query perform well in our experiments. The simplicity also enables us to feed the same features to vanilla and sequential DPPs for fair comparison.



\vspace{-8pt}

\subsection{Evaluation} \label{sEvaluation}
We evaluate a system generated video summary by contrasting it against the ``groundtruth'' summary. The comparison is based on the dense text annotations~\cite{videoSET}. In particular, the video summaries are mapped to text representations and then compared by the ROUGE-SU metric~\cite{lin2004rouge}. We report the precision, recall, F-measure returned by ROUGE-SU. 

In addition, we also introduce a new metric, called \textit{hitting recall}, to evaluate the system summaries from the query-focused perspective. Given the input query $q$ and long video ${\cal Y}$, denote by $S^q$ the shots relevant to the query in the ``groundtruth'' summary, and $S^q_\textsc{system}$ the query-relevant shots hit by a video summarizer. The hitting recall is calculated by $\textsc{hr}=|S^q_\textsc{system}|/|S^q|$, where $|\cdot|$ is the cardinality of a set. For our SH-DPP model, we report the hitting recall for both the overall summaries and those by the $Z$-layer only.

\vspace{-8pt}

\begin{table}[h]
	\vspace{-20pt}
	\centering
	\caption{Results of query-focused video summarization with
		{\bf bi-concept} queries.}\label{tTwo}
	
	\begin{tabular}{c|c|c|c|c|c||c|c|c|c|c}
		\hline \toprule
		Patient Users & \multicolumn{5}{c||}{UTE (\%)} & \multicolumn{5}{c}{TV episodes (\%)}\tabularnewline
		\hline 
		& F & Prec. & Recall & \textsc{hr} & \textsc{hr}$_Z$ & F & Prec. & Recall & \textsc{hr} & \textsc{hr}$_Z$\tabularnewline
		\hline 
		Sampling &\textbf{22.12}&\textbf{35.07}&17.11&23.61 & n/a & 27.99&34.75&24.36&16.00 & n/a\tabularnewline
		\hline 
		Ranking & 20.66 & 24.35 & 18.38 & 22.05 & n/a & 32.19&39.96&32.19&16.61 & n/a\tabularnewline
		\hline
		SubMod ~\cite{gygli2015video} & 20.98 & 31.40 & 26.99 & 30.10 & n/a & 32.19 & \textbf{41.59} & 27.01 & 21.69 & n/a\tabularnewline
		\hline
		Quasi ~\cite{zhao2014quasi} & 12.45 & 19.47 & 13.14 & 14.95 & n/a & 31.88 & 27.49 & 41.69 & 19.67 & n/a\tabularnewline
		\hline
		DPP ~\cite{kulesza2012learning} & 15.7&19.22&32.08&30.94 & n/a & 29.62&35.26&34.00&21.29 & n/a\tabularnewline
		\hline 
		seqDPP ~\cite{gong2014diverse}& 18.85&20.59&35.83&31.91 & n/a & 27.96&23.80&35.62&14.08 & n/a\tabularnewline
		\hline 
		SH-DPP (ours) & 21.27 & 17.87 & \textbf{41.65} & \textbf{38.26} & \textbf{36.92} & \textbf{37.02} & 38.41 & \textbf{36.82} & \textbf{23.76} & 20.35 \tabularnewline
		\hline 
		\hline \toprule
		Impatient Users & \multicolumn{5}{c||}{UTE (\%)} & \multicolumn{5}{c}{TV episodes (\%)}\tabularnewline
		\hline 
		& F & P & R & \textsc{hr} & \textsc{hr}$_Z$ & F & P & R & \textsc{hr} & \textsc{hr}$_Z$\tabularnewline
		\hline 
		Sampling & 25.44&44.16&18&6.48 & n/a & 33.74&\textbf{41.03}&28.8&13.03& n/a\tabularnewline
		\hline 
		Ranking & 17.92&21.86&15.46&4.4 & n/a & 29.67&37.56&24.72&15.43 & n/a\tabularnewline
		\hline 
		SubMod ~\cite{gygli2015video} & \textbf{27.10} & \textbf{51.79} & 18.85 & 8.05 & n/a & 29.41 & 38.51 & 23.85 & 8.65 & n/a\tabularnewline
		\hline
		Quasi ~\cite{zhao2014quasi}& 11.52 & 42.32 & 7.06 & 1.63 & n/a & 25.09 & 27.25 & 23.71 & 17.06 & n/a\tabularnewline
		\hline
		DPP ~\cite{kulesza2012learning} & 14.36&30.9&16.18&12.54 & n/a & 26.01&28.85&39.15&\textbf{18.86} & n/a\tabularnewline
		\hline 
		seqDPP ~\cite{gong2014diverse}& 12.93&7.89&43.39&12.68 & n/a & 23.35&16.60&39.69&12.56 & n/a\tabularnewline
		\hline 
		SH-DPP (ours) & 25.56&18.51&\textbf{45.21}&\textbf{22.91}&11.57 & \textbf{35.36}&30.94&\textbf{42.02}&17.07&17.07\tabularnewline
		\hline 
	\end{tabular}
\end{table}
\vspace{-8pt}
\subsection{Implementation Details}
Here we report some details in our implementation of SH-DPP. Out of the four videos in either UTE or TV, we use three videos  for training and the remaining one for testing. Each video is taken for testing once and then the averaged  results are reported. In the training phase, there are two hyper-parameters in our approach: $\lambda_1$ and $\lambda_2$ (cf.\ eq.~(\ref{eMLE})). We choose their values by the leave-one-video-out strategy (over the training videos only). The lower-dimensions of $\mat{W}$ and $\mat{V}$ are both fixed to 10, the same as used in seqDPP~\cite{gong2014diverse}. Varying this number has little effect to the results and we leave the study about it to the Supplementary Material. We put 10 shots in the ground set $\mathcal{Y}_t$ at each time step, and also examine the ground sets of the other sizes in the experiments. 

We train our model SH-DPP using bi-concept queries. However, we test it using not only the bi-concept queries but also novel three-concept queries.

\section{Experimental Results}
\vspace{-4pt}
This section presents the comparison results of our approach and some competitive baselines,  effect of the ground set size, and finally  qualitative results.

\subsection{Comparison Results}

\begin{wraptable}{r}{7.85cm}
\small
\vspace{-12pt}
	\caption{Results of query-focused video summarization with novel
		{\bf three-concept} queries. } \label{tThree}
		\vspace{-8pt}
	\begin{tabular}{c|c|c|c||c|c|c}
		\hline \toprule
		Patient Users & \multicolumn{3}{c||}{UTE (\%)} & \multicolumn{3}{c}{TV episodes (\%)}\tabularnewline
		\hline 
		& F  & \textsc{hr} & \textsc{hr}$_Z$ & F &  \textsc{hr} & \textsc{hr}$_Z$\tabularnewline
		\hline 
		DPP & 20.7 &38.53 & n/a & 29.84 & 21.68 & n/a\tabularnewline
		\hline 
		seqDPP & 18.03 & 30.3 & n/a & 24.29 & 14.15 & n/a\tabularnewline
		\hline 
		SH-DPP (ours) & \textbf{24.54} & \textbf{41.23} & 40.43 & \textbf{36.3} & \textbf{24.73} & 21.71 \tabularnewline
		\hline 
		\hline \toprule
		Impatient Users & \multicolumn{3}{c||}{UTE (\%)} & \multicolumn{3}{c}{TV episodes (\%)}\tabularnewline
		\hline 
		& F & \textsc{hr} & \textsc{hr}$_Z$ & F & \textsc{hr} & \textsc{hr}$_Z$\tabularnewline
		\hline 
		DPP & 14.77 & 17.28 & n/a & 24.71  &\textbf{18.31} & n/a\tabularnewline
		\hline 
		seqDPP & 19.4  & 19.17 & n/a & 29.31 & 10.09 & n/a\tabularnewline
		\hline 
		SH-DPP (ours) & \textbf{29.59} & \textbf{25.82} & 15.36 & \textbf{33.94} & 17.39 & 12.33 \\
		\hline 
	\end{tabular}
	\vspace{-18pt}
\end{wraptable} 

\vspace{-4pt}
Table~\ref{tTwo} shows the results of different summarizers for the query-focused video summarization when the patient and impatient users supply {\bf bi-concept} queries, while Table~\ref{tThree} includes the results for novel {\bf three-concept} queries. Note that only  bi-concept queries are used to train the summarizers. We report the results on both UTE and TV datasets, and contrast our SH-DPP to the following methods: 1) uniformly sampling $K$ shots, 2) ranking, where for each query we apply the corresponding concept detectors to the shots, assign to a shot a ranking score as the maximum detection score, and then keep the top $K$ shots, 3) vanilla DPP~\cite{kulesza2012learning}, where we remove the dependency between adjacent subset selection variables in Figure~\ref{fig:SOSeqDPP}(a),  4) seqDPP~\cite{gong2014diverse}, 5) SubMod~\cite{gygli2015video}, where convex combination of a set of objectives is learned from user summaries, and 6) Quasi ~\cite{zhao2014quasi} which is an unsupervised method based on group sparse coding. We let $K$ be the number of shots in the groundtruth summary; therefore, such privileged information makes 1), 2), and 5) actually strong baselines. We use the same ground sets, whose sizes are fixed to 10 and are studied in Section~\ref{sSize}, for DPP, seqDPP, and our SH-DPP. All the results are evaluated by the F-measure, Precision, and Recall of ROUGE-SU, as well as the hitting recall (HR) (cf.\ Section~\ref{sEvaluation}). 


Interesting insights can be inferred from Tables \ref{tTwo} and \ref{tThree}. An immediate observation is that our SH-DPP is able to generate better overall summaries as our average F-measure scores are higher than the others'. Furthermore, our method is able to adapt itself to two essentially different datasets, the UTE daily life egocentric videos and TV episodes. 

On UTE, we expect both SH-DPP and seqDPP to outperform vanilla DPP, because the egocentric videos  are very long and include many unwanted scenes, and thus the dependency between different subset selection variables  helps eliminate repetitions. In contrast, as mentioned in \ref{subsec:Dataset}, the TV episodes are from the world of professional recording, and the scenes rapidly change from shots to shots. Therefore, in this case, the dependency is weak and DPP may be able to catch up seqDPP's performance. These hypotheses are verified in the results, if we compare DPP with seqDPP in Tables  \ref{tTwo} and \ref{tThree}.

Another important observation is that in 6 out of the 8 experiments: \{patient and impatient users\} on \{UTE and TV datasets\} by \{bi-concept and novel three-concept queries\}, the proposed SH-DPP has better hitting recalls than the other methods, indicating a better response to the user queries. Moreover, the hitting recalls are mainly captured by the $Z$-layer---the columns HR$_Z$ are the hitting recalls of the shots selected by the $Z$-layer only of SH-DPP.

\begin{wrapfigure}{r}{0.5\textwidth}
	\centering
		\vspace{-20pt}
	\includegraphics[width=0.48\textwidth]{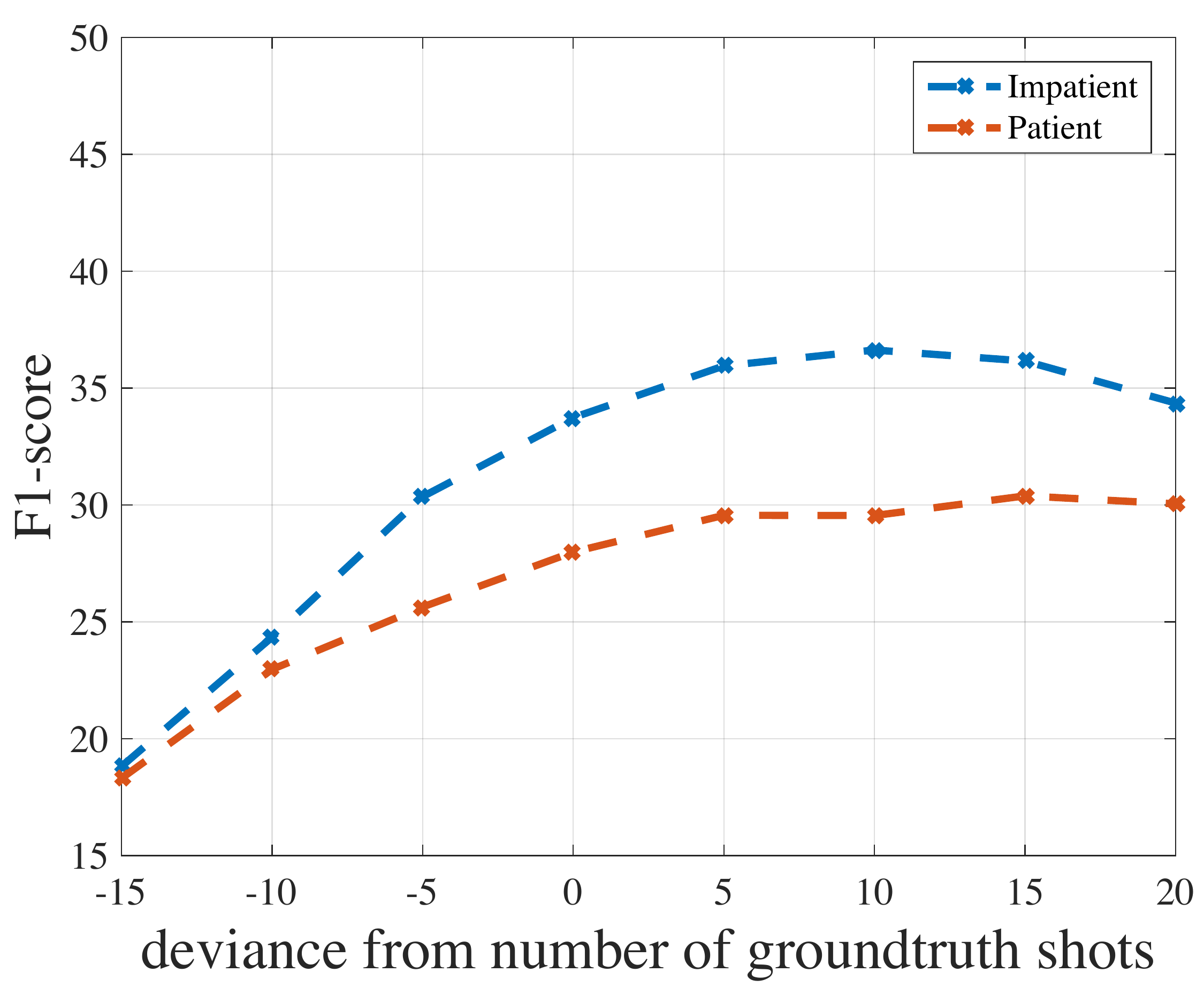}
		\vspace{-3pt}
	\caption{The effect of number of selected shots on the performance of uniform sampling. }
	\label{fig:dev}
	\vspace{-15pt}
\end{wrapfigure}

As it can be noticed from Table ~\ref{tTwo} uniform sampling have a competitive performance compared to baselines and even outperforms SH-DPP in one scenario. The relatively good performance of random sampling can be explained by looking into the evaluation metric. ROUGE essentially evaluates the summaries by common word/phrase count and penalizing long or short summaries. Thus, accessing the number of groundtruth shots gives an advantage to Sampling. Figure ~\ref{fig:dev} illustrates the change of performance when we deviate from the number of shots in groundtruth summary. This figure was generated using TV Episodes dataset for both patient and impatient user cases. 

\vspace{-6pt}

\subsection{A Peek into the SH-DPP Summarizer}
\begin{figure}[t]
\centering
		\includegraphics[scale=0.25]{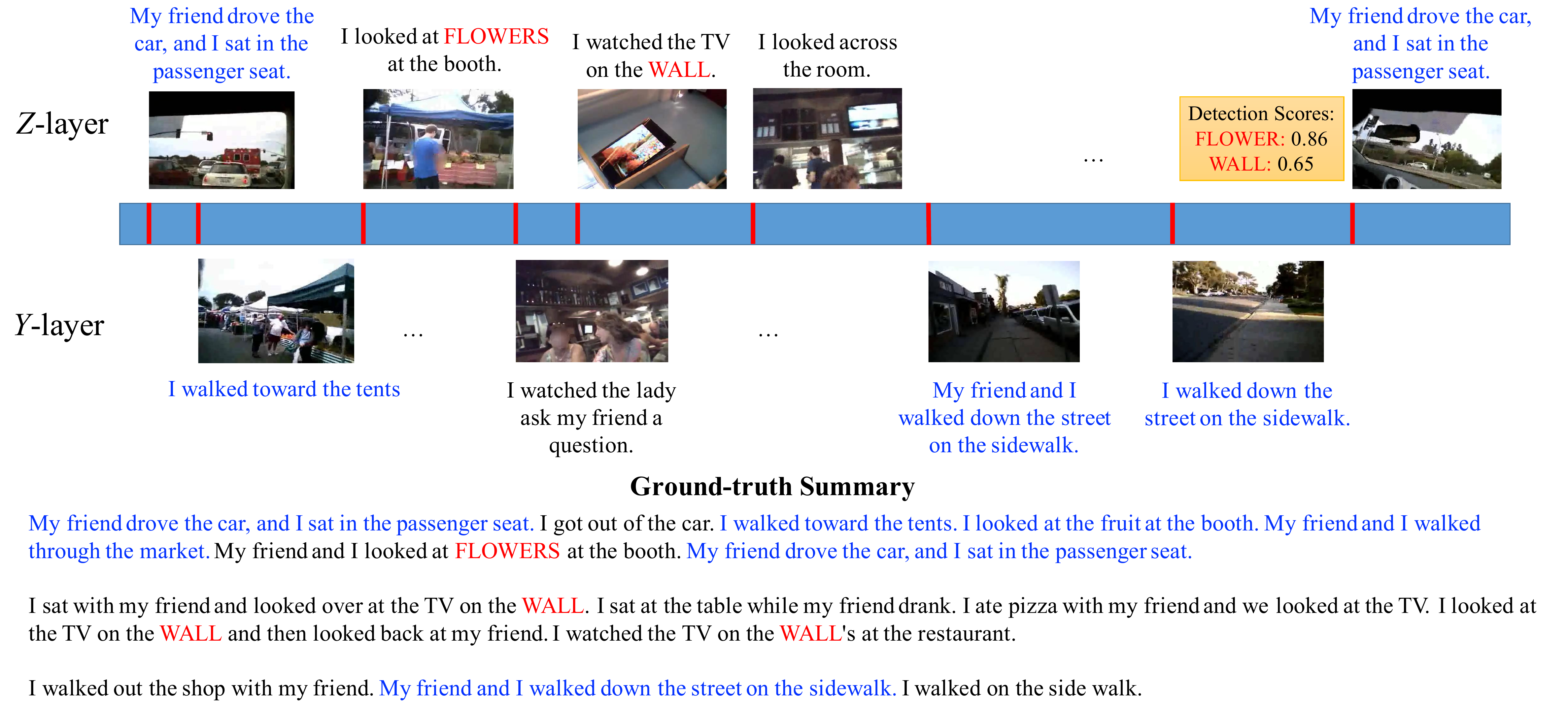}
	\caption{A peek into SH-DPP. Given the query \textcolor{red}{\textsc{flowers}}+\textcolor{red}{\textsc{wall}}, the $Z$-layer of SH-DPP is supposed to summarize the shots relevant to the query. Conditioning on those results, the $Y$-layer summarizes the remaining video.} \label{fQuality}
	\vspace{-10pt}
\end{figure}

Figure~\ref{fQuality} is an exemplar summary for the query  \textcolor{red}{\textsc{flowers}}+\textcolor{red}{\textsc{wall}} by SH-DPP. For each shot in the summary, we show the middle frame of that shot and the corresponding textual description. The groundtruth summary is also included at the bottom half of the figure. We can see that some query-relevant shots are successfully selected by the $Z$-layer. Conditioning on those, the $Y$-layer summarizes the remaining video. We highlight the text descriptions (in the \textcolor{blue}{blue} color) that have exact matches in the groundtruth. However, please note that the other sentences are also highly correlated with some groundtruth sentences, for instance, \textit{``I looked at flowers at the booth''} selected by the $Z$-layer versus \textit{``my friend and I looked at flowers at the booth''} in the groundtruth summary. 

One may wonder why the top-right shot is selected by the $Z$-layer, since it is visually not relevant to either \textcolor{red}{\textsc{flowers}} or \textcolor{red}{\textsc{wall}}. Inspection tells that it is due to the failure of the concept detectors;  the detection scores are 0.86 and 0.65 out of 1 for \textcolor{red}{\textsc{flowers}} and \textcolor{red}{\textsc{wall}}, respectively. We may improve our SH-DPP for the query-focused video summarization by using better concept detectors. 
\vspace{-6pt}

\begin{wrapfigure}{r}{0.5\textwidth}
	\centering
		\vspace{-20pt}
	\includegraphics[width=0.48\textwidth]{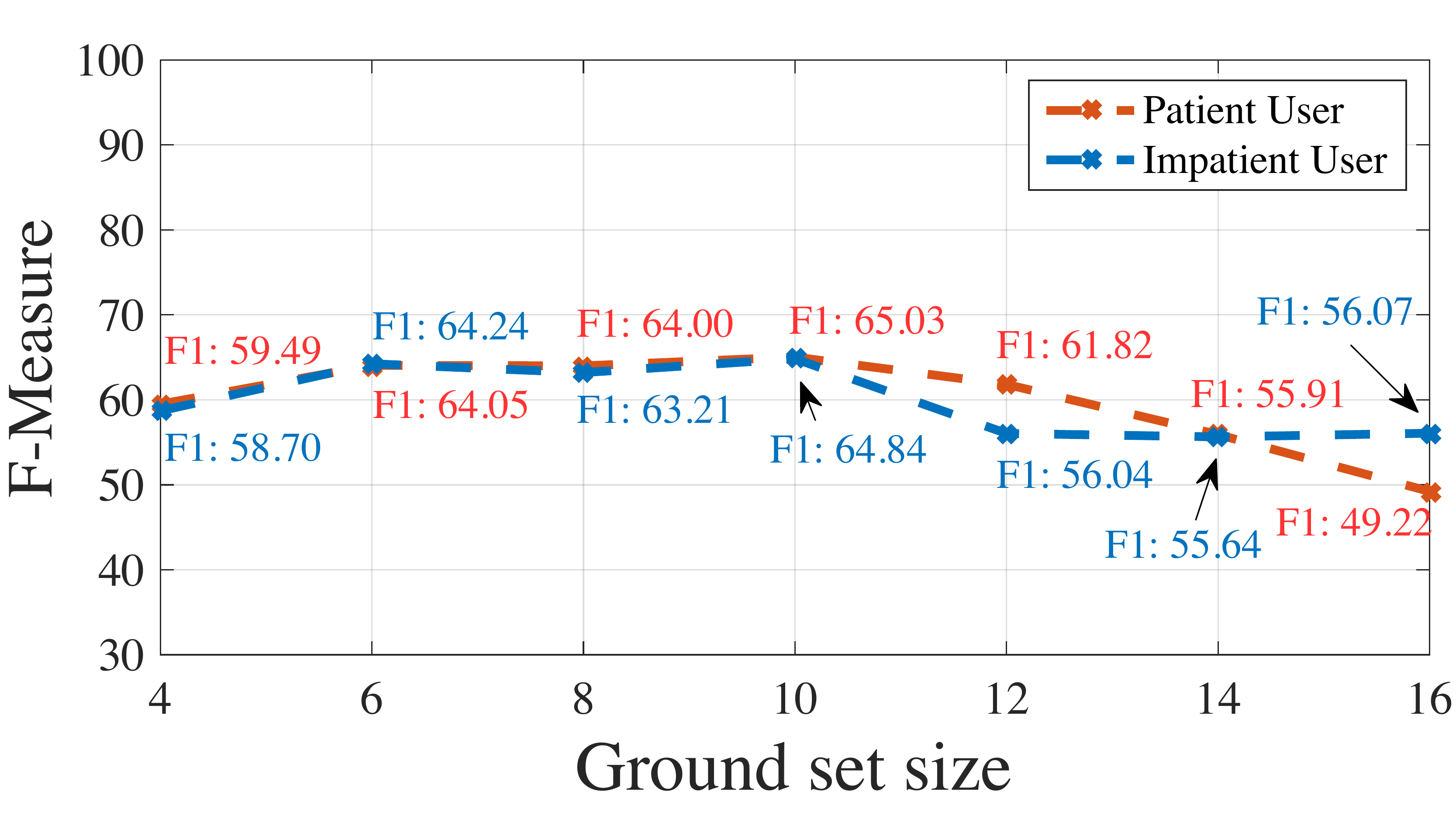}
		\vspace{-3pt}
	\caption{The effect of ground set size on the performance of SH-DPP. }
	\label{fig:GSEff}
	\vspace{-15pt}
\end{wrapfigure}

\subsection{Effect of the ground set size} \label{sSize}

In this section, we examine how changing the ground set size affects the performance of the proposed method. To generate Figure \ref{fig:GSEff}, we train our SH-DPP with different ground set sizes varying from 4 to 16 by a step size of 2, and then evaluate them on the test data. The results are obtained on the TV episodes where each shot lasts for 10 seconds. 
We can see that the F-measure scores only change slightly for the ground sets of 6 to 10 shots. The results decrease a little for very small ground sets (4 shots, 40s) or large sets (16 shots, 160s). This is probably because either too small or too large ground sets fail to follow the internal short memories of the annotators, 
and further validates our Markov assumption for the video summarization---when users summarize very long videos, they tend to use short memories to impose only local diversities to the selected shots.

\subsubsection{In the Supplementary Material:} We describe the detailed training algorithm for SH-DPP using gradient descent. The resultant optimization problem is non-convex; we explain how to choose the initializations by the leaving-one-video-out strategy. We also show that the SH-DPP results remain stable for different lower-dimensions of $\mat{W}$ and $\mat{V}$. Finally, more qualitative results are included in the Supplementary Material. 

		\vspace{-4pt}


\section{Conclusions}
In this paper, we examined a query-focused video summarization problem, in which the decision to select a video shot to the summary depends on both 1) the relevance between the shot and the query and 2) the importance of the shot in the context of the video. To tackle this problem, we developed a probabilistic model, Sequential and Hierarchical Determinantal Point Process (SH-DPP), as well as efficient learning and inference algorithms for it. Our SH-DPP summarizer can conveniently handle extremely long videos or online streaming videos. On two benchmark datasets for video summarization, our approach significantly outperforms some competing baselines. To the best of our knowledge, ours is the first work on query-focused video summarization, and has a great potential to be used in search engines, e.g., to display snippets of videos.

\clearpage

\bibliographystyle{splncs}
\bibliography{vsumm}

\begin{thebibliography}{10}

\bibitem{pritch2007webcam}
Pritch, Y., Rav-Acha, A., Gutman, A., Peleg, S.:
\newblock Webcam synopsis: Peeking around the world.
\newblock In: Computer Vision, 2007. ICCV 2007. IEEE 11th International
  Conference on, IEEE (2007)  1--8

\bibitem{pal2005interactive}
Pal, C., Jojic, N.:
\newblock Interactive montages of sprites for indexing and summarizing security
  video.
\newblock In: CVPR, 2005. CVPR 2005. IEEE Computer Society Conference on.
  Volume~2., IEEE (2005)

\bibitem{kang2006space}
Kang, H.W., Matsushita, Y., Tang, X., Chen, X.Q.:
\newblock Space-time video montage.
\newblock In: CVPR, 2006 IEEE computer society conference on. Volume~2., IEEE
  (2006)

\bibitem{jiang2009advances}
Jiang, R.M., Sadka, A.H., Crookes, D.:
\newblock Advances in video summarization and skimming.
\newblock In: Recent Advances in Multimedia Signal Processing and
  Communications.
\newblock Springer (2009)  27--50

\bibitem{rav2006making}
Rav-Acha, A., Pritch, Y., Peleg, S.:
\newblock Making a long video short: Dynamic video synopsis.
\newblock In: CVPR, 2006 IEEE Computer Society Conference on. Volume~1., IEEE
  (2006)

\bibitem{goldman2006schematic}
Goldman, D.B., Curless, B., Salesin, D., Seitz, S.M.:
\newblock Schematic storyboarding for video visualization and editing.
\newblock In: ACM Transactions on Graphics (TOG). Volume~25., ACM (2006)
  862--871

\bibitem{liu2002optimization}
Liu, T., Kender, J.R.:
\newblock Optimization algorithms for the selection of key frame sequences of
  variable length.
\newblock In: Computer VisionECCV 2002.
\newblock Springer (2002)  403--417

\bibitem{aner2002video}
Aner, A., Kender, J.R.:
\newblock Video summaries through mosaic-based shot and scene clustering.
\newblock In: Computer VisionECCV 2002.
\newblock Springer (2002)

\bibitem{vasconcelos1998spatiotemporal}
Vasconcelos, N., Lippman, A.:
\newblock A spatiotemporal motion model for video summarization.
\newblock In: CVPR, 1998. Proceedings. 1998 IEEE Computer Society Conference
  on, IEEE (1998)  361--366

\bibitem{wolf1996key}
Wolf, W.:
\newblock Key frame selection by motion analysis.
\newblock In: Acoustics, Speech, and Signal Processing, 1996. ICASSP-96.
  Conference Proceedings., 1996 IEEE International Conference on. Volume~2.,
  IEEE (1996)  1228--1231

\bibitem{lee2012unified}
Lee, K.M., Kwon, J.:
\newblock A unified framework for event summarization and rare event detection.
\newblock In: 2012 IEEE Conference on CVPR, IEEE (2012)

\bibitem{cong2012towards}
Cong, Y., Yuan, J., Luo, J.:
\newblock Towards scalable summarization of consumer videos via sparse
  dictionary selection.
\newblock Multimedia, IEEE Transactions on \textbf{14}(1) (2012)  66--75

\bibitem{ngo2003automatic}
Ngo, C., Ma, Y., Zhang, H.:
\newblock Automatic video summarization by graph modeling.
\newblock In: Computer Vision, 2003. Proceedings. Ninth IEEE International
  Conference on, IEEE (2003)

\bibitem{khosla2013large}
Khosla, A., Hamid, R., Lin, C.J., Sundaresan, N.:
\newblock Large-scale video summarization using web-image priors.
\newblock In: Proceedings of the IEEE Conference on CVPR. (2013)

\bibitem{kim2014joint}
Kim, G., Sigal, L., Xing, E.:
\newblock Joint summarization of large-scale collections of web images and
  videos for storyline reconstruction.
\newblock In: Proceedings of the IEEE Conference on CVPR. (2014)

\bibitem{xiong2014detecting}
Xiong, B., Grauman, K.:
\newblock Detecting snap points in egocentric video with a web photo prior.
\newblock In: Computer Vision--ECCV 2014.
\newblock Springer (2014)  282--298

\bibitem{chu2015video}
Chu, W.S., Song, Y., Jaimes, A.:
\newblock Video co-summarization: Video summarization by visual co-occurrence.
\newblock In: Proceedings of the IEEE Conference on CVPR. (2015)

\bibitem{song2015tvsum}
Song, Y., Vallmitjana, J., Stent, A., Jaimes, A.:
\newblock Tvsum: Summarizing web videos using titles.
\newblock In: Proceedings of the IEEE Conference on CVPR. (2015)

\bibitem{liu2015multi}
Liu, W., Mei, T., Zhang, Y., Che, C., Luo, J.:
\newblock Multi-task deep visual-semantic embedding for video thumbnail
  selection.
\newblock In: Proceedings of the IEEE Conference on CVPR. (2015)

\bibitem{potapov2014category}
Potapov, D., Douze, M., Harchaoui, Z., Schmid, C.:
\newblock Category-specific video summarization.
\newblock In: Computer Vision--ECCV 2014.
\newblock Springer (2014)

\bibitem{sun2014ranking}
Sun, M., Farhadi, A., Seitz, S.:
\newblock Ranking domain-specific highlights by analyzing edited videos.
\newblock In: Computer Vision--ECCV 2014.
\newblock Springer (2014)

\bibitem{xu2015gaze}
Xu, J., Mukherjee, L., Li, Y., Warner, J., Rehg, J.M., Singh, V.:
\newblock Gaze-enabled egocentric video summarization via constrained
  submodular maximization.
\newblock In: Proceedings of the IEEE Conference on CVPR. (2015)

\bibitem{gygli2014creating}
Gygli, M., Grabner, H., Riemenschneider, H., Van~Gool, L.:
\newblock Creating summaries from user videos.
\newblock In: Computer Vision--ECCV 2014.
\newblock Springer (2014)

\bibitem{lu2013story}
Lu, Z., Grauman, K.:
\newblock Story-driven summarization for egocentric video.
\newblock In: Proceedings of the IEEE Conference on CVPR. (2013)

\bibitem{lee2015predicting}
Lee, Y.J., Grauman, K.:
\newblock Predicting important objects for egocentric video summarization.
\newblock International Journal of Computer Vision \textbf{114}(1) (2015)
  38--55

\bibitem{liu2010hierarchical}
Liu, D., Hua, G., Chen, T.:
\newblock A hierarchical visual model for video object summarization.
\newblock Pattern Analysis and Machine Intelligence, IEEE Transactions on
  \textbf{32}(12) (2010)  2178--2190

\bibitem{gong2014diverse}
Gong, B., Chao, W.L., Grauman, K., Sha, F.:
\newblock Diverse sequential subset selection for supervised video
  summarization.
\newblock In: Advances in Neural Information Processing Systems. (2014)
  2069--2077

\bibitem{gygli2015video}
Gygli, M., Grabner, H., Van~Gool, L.:
\newblock Video summarization by learning submodular mixtures of objectives.
\newblock In: Proceedings of the IEEE Conference on CVPR. (2015)

\bibitem{nenkova2012survey}
Nenkova, A., McKeown, K.:
\newblock A survey of text summarization techniques.
\newblock In: Mining Text Data.
\newblock Springer (2012)  43--76

\bibitem{kulesza2012determinantal}
Kulesza, A., Taskar, B.:
\newblock Determinantal point processes for machine learning.
\newblock arXiv preprint arXiv:1207.6083 (2012)

\bibitem{ghosh2012discovering}
Ghosh, J., Lee, Y.J., Grauman, K.:
\newblock Discovering important people and objects for egocentric video
  summarization.
\newblock In: 2012 IEEE Conference on CVPR, IEEE (2012)

\bibitem{videoSET}
Yeung, S., Fathi, A., Fei-Fei, L.:
\newblock Videoset: Video summary evaluation through text.
\newblock arXiv preprint arXiv:1406.5824 (2014)

\bibitem{daume2006bayesian}
Daum{\'e}~III, H., Marcu, D.:
\newblock Bayesian query-focused summarization.
\newblock In: Proceedings of the 21st International Conference on Computational
  Linguistics and the 44th annual meeting of the Association for Computational
  Linguistics, Association for Computational Linguistics (2006)

\bibitem{schilder2008fastsum}
Schilder, F., Kondadadi, R.:
\newblock Fastsum: fast and accurate query-based multi-document summarization.
\newblock In: Proceedings of the 46th Annual Meeting of the Association for
  Computational Linguistics on Human Language Technologies: Short Papers,
  Association for Computational Linguistics (2008)  205--208

\bibitem{gupta2007measuring}
Gupta, S., Nenkova, A., Jurafsky, D.:
\newblock Measuring importance and query relevance in topic-focused
  multi-document summarization.
\newblock In: Proceedings of the 45th Annual Meeting of the ACL on Interactive
  Poster and Demonstration Sessions, Association for Computational Linguistics
  (2007)  193--196

\bibitem{ellouze2010s}
Ellouze, M., Boujemaa, N., Alimi, A.M.:
\newblock Im (s) 2: Interactive movie summarization system.
\newblock Journal of Visual Communication and Image Representation
  \textbf{21}(4) (2010)  283--294

\bibitem{xiong2015storyline}
Xiong, B., Kim, G., Sigal, L.:
\newblock Storyline representation of egocentric videos with an applications to
  story-based search.
\newblock In: Proceedings of the IEEE International CVPR. (2015)

\bibitem{kulesza2012learning}
Kulesza, A., Taskar, B.:
\newblock Learning determinantal point processes.
\newblock arXiv preprint arXiv:1202.3738 (2012)

\bibitem{chaolarge}
Chao, W.L., Gong, B., Grauman, K., Sha, F.:
\newblock Large-margin determinantal point processes

\bibitem{affandi2012markov}
Affandi, R.H., Kulesza, A., Fox, E.B.:
\newblock Markov determinantal point processes.
\newblock arXiv preprint arXiv:1210.4850 (2012)

\bibitem{borth2013sentibank}
Borth, D., Chen, T., Ji, R., Chang, S.F.:
\newblock Sentibank: large-scale ontology and classifiers for detecting
  sentiment and emotions in visual content.
\newblock In: Proceedings of the 21st ACM international conference on
  Multimedia, ACM (2013)

\bibitem{oliva2001modeling}
Oliva, A., Torralba, A.:
\newblock Modeling the shape of the scene: A holistic representation of the
  spatial envelope.
\newblock International journal of computer vision \textbf{42}(3) (2001)
  145--175

\bibitem{ojala2002multiresolution}
Ojala, T., Pietik{\"a}inen, M., M{\"a}enp{\"a}{\"a}, T.:
\newblock Multiresolution gray-scale and rotation invariant texture
  classification with local binary patterns.
\newblock Pattern Analysis and Machine Intelligence, IEEE Transactions on
  \textbf{24}(7) (2002)  971--987

\bibitem{yu2013designing}
Yu, F., Cao, L., Feris, R., Smith, J., Chang, S.F.:
\newblock Designing category-level attributes for discriminative visual
  recognition.
\newblock In: Proceedings of the IEEE Conference on CVPR. (2013)

\bibitem{lin2004rouge}
Lin, C.Y.:
\newblock Rouge: A package for automatic evaluation of summaries.
\newblock In: Text summarization branches out: Proceedings of the ACL-04
  workshop. Volume~8. (2004)

\bibitem{zhao2014quasi}
Zhao, B., Xing, E.:
\newblock Quasi real-time summarization for consumer videos.
\newblock In: Proceedings of the IEEE Conference on CVPR. (2014)

\end{thebibliography}
\end{document}